%% file: main.tex
\documentclass{article}

\usepackage{times}

\usepackage[margin=1in]{geometry}

\usepackage[round,authoryear]{natbib}

\input{preamble.tex}

\title{\textsf{Fora}: From Weight-Space to Function-Space Protection\\ in Capability-Preserving Fine-Tuning}

\author{
  Rui Zhou\thanks{Correspondence: \url{zzz_rrr@mail.dlut.edu.cn}} \\
  \textit{Dalian University of Technology} \\
  \and
  Tianci Xie\thanks{\url{xietianci2004@gmail.com}} \\
  \textit{Zhejiang University}
}

\begin{document}

\maketitle

\begin{abstract}
\input{sec/0_abstract.tex}
\end{abstract}

\input{sec/1_intro.tex}
\input{sec/2_related_work.tex}
\input{sec/3_method.tex}
\input{sec/5_experiments.tex}
\input{sec/6_discussion.tex}
\input{sec/7_limitations.tex}
\input{sec/8_conclusion.tex}

\newpage
\bibliographystyle{plainnat}
\bibliography{references}

\newpage
\appendix
\input{appendix/A_implementation.tex}
\input{appendix/B_per_seed.tex}
\input{appendix/C_full_peft.tex}
\input{appendix/D_spectral_only.tex}
\input{appendix/E_extensions.tex}

\end{document}

%% file: preamble.tex
\usepackage{amsmath,amssymb,amsthm}
\usepackage{bm}

\newtheorem{proposition}{Proposition}

\theoremstyle{plain}

\usepackage{graphicx}
\usepackage{tikz}
\usetikzlibrary{shapes,arrows,positioning,fit,backgrounds,calc,decorations.pathreplacing}
\usepackage{caption}
\usepackage{subcaption}

\usepackage{booktabs}
\usepackage{multirow}
\usepackage{array}
\usepackage{makecell}

\usepackage{enumitem}
\usepackage{xspace}
\usepackage{microtype}
\usepackage{url}
\usepackage[hidelinks]{hyperref}

\newcommand{\W}{\mathbf{W}}
\newcommand{\dW}{\Delta\!\W}
\newcommand{\hW}{\widehat{\W}}
\newcommand{\M}{\mathbf{M}}
\newcommand{\Sres}{\mathbf{S}_{\mathrm{res}}}
\newcommand{\A}{\mathcal{A}}
\newcommand{\B}{\mathbf{B}}

\newcommand{\C}{\mathbf{C}}
\newcommand{\D}{\mathbf{D}}
\newcommand{\X}{\mathbf{X}}
\newcommand{\U}{\mathbf{U}}
\newcommand{\V}{\mathbf{V}}
\newcommand{\Q}{\mathbf{Q}}
\newcommand{\I}{\mathbf{I}}
\newcommand{\bdelta}{\bm{\delta}}

\newcommand{\R}{\mathbb{R}}

\newcommand{\PU}{P_{\!\U}}
\newcommand{\PV}{P_{\!\V}}
\newcommand{\PQ}{P_{\!\Q}}

\newcommand{\calspan}{\operatorname{span}}
\newcommand{\diag}{\operatorname{diag}}

\newcommand{\Dpres}{\mathcal{D}_{\text{pres}}}
\newcommand{\Dnew}{\mathcal{D}_{\text{new}}}

\newcommand{\Wzero}{\W_0}

\newcommand{\Fora}{\textsf{Fora}}
\newcommand{\Foralr}{\textsf{Fora}-LR}
\newcommand{\FProj}{\textsf{FProj}}
\newcommand{\FProjlr}{\textsf{FProj}-LR}
\newcommand{\WProj}{\textsf{WProj}}
\newcommand{\FullFT}{Full FT}
\newcommand{\EWC}{EWC}
\newcommand{\LtwoSP}{L2-SP}
\newcommand{\LwF}{LwF}

%% file: sec/0_abstract.tex

Full fine-tuning adapts large language models to new tasks but can erode capabilities they already possess. Existing remedies protect through \emph{proxies}---parameter distances, importance penalties, output matching, or the dominant singular directions of the weights---none of which answers the operative question: \emph{which activation directions does the preserved capability rely on?} We argue a capability is characterized more faithfully by the activation subspace it induces than by the singular geometry of the weight matrix, and on this view develop \textbf{function-space protection}, instantiated as \Fora{} (\emph{Function-space Orthogonal Residual Adaptation}). From label-free calibration inputs we estimate, per layer, the principal directions $\Q$ of the input-activation covariance and form a right projector $\PQ = \I - \Q\Q^\top$; paired with a left projector $\PU$ from the weight's SVD, the \Fora{} update is $\dW = \PU\M\PQ + \U_2\D_\delta\V_2^\top$---a high-capacity branch structurally barred from reading the capability's function directions, plus a narrow spectral channel for controlled plasticity. The construction extends to parameter-efficient adaptation via $\M \to \tfrac{\alpha}{r}\B\A$. Across three settings on Qwen3-1.7B (COGS and GSM8K learned while preserving translation, translation learned while preserving math), it consistently outperforms weight-space projection and standard regularization on preservation, with only a small new-task trade-off in the math-preservation setting. A controlled ablation isolating the projection \emph{source} shows the advantage comes not from projection itself, but from projecting onto capability-derived rather than weight-derived directions. Code is available at \url{https://github.com/zrui239/FORA}.

%% file: sec/1_intro.tex

\section{Introduction}
\label{sec:intro}

Full fine-tuning---updating all parameters of a pre-trained model~\citep{vaswani2017attention} under unconstrained gradients---remains the most expressive way to adapt a large language model to a new task~\citep{han2024peftsurvey}. This expressiveness is also a liability: unconstrained gradients are free to overwrite the functional pathways of capabilities the model already possesses. A model expected to retain translation quality while learning to parse semantics, or to retain arithmetic ability while learning a new language pair, must avoid this drift, yet the standard fine-tuning objective does nothing to discourage it.

A long line of work mitigates such forgetting~\citep{mccloskey1989catastrophic,french1999catastrophic,parisi2019continual,shi2024clsurvey}. Regularization methods anchor the weights through importance-weighted~\citep{kirkpatrick2017overcoming} or plain $\ell_2$~\citep{li2018explicit} penalties; distillation methods keep the output distribution close to the original model's~\citep{li2017learning}; gradient-projection methods steer updates away from stored task subspaces~\citep{saha2021gradient,wang2024oplora}; and weight-decomposition methods restrict adaptation to chosen singular components of the pre-trained weight~\citep{meng2024pissa,liu2024dora}. Each of these protects a capability through a \emph{proxy}: parameter importance, output behavior, gradient history, or weight geometry. None of them directly encodes the set of activation directions that the capability uses during inference.

\paragraph{Our claim.} \textbf{A preserved capability is better characterized by the activation subspace it induces in each layer than by the dominant singular directions of the weight matrix.} The top right-singular vectors $\V_1$ of a weight $\Wzero$ summarize where that weight is \emph{large}---an aggregate over everything the model has ever learned. They need not align with where a \emph{specific} capability is functionally active: translation may rely on activation patterns spread across many moderate singular components, or oriented diagonally with respect to the SVD basis. Protecting $\V_1$ therefore constrains a subspace that is largely misaligned with the capability.

\paragraph{Method.} We address this by deriving the protected subspace from the capability itself. Forwarding label-free calibration inputs of the preserved capability, we collect per-layer input activations, take the top-$k$ eigenvectors $\Q$ of their covariance, and form a \emph{function-space} right projector $\PQ = \I - \Q\Q^\top$. The full-rank update is
\begin{equation*}
\dW \;=\; \underbrace{\PU\,\M\,\PQ}_{\text{high-capacity, function-protected}} \;+\; \underbrace{\U_2\D_\delta\V_2^\top}_{\text{controlled spectral channel}},
\end{equation*}
where $\PU = \I - \U_1\U_1^\top$ prevents the high-capacity branch from writing the principal output directions of $\Wzero$, and $\PQ$ prevents it from reading the capability's function directions. The two channels play complementary roles: on inputs aligned with the protected capability the dense branch contributes nothing and the response is determined solely by the narrow spectral channel---a small, calibrated adjustment rather than an outright freeze---while on all other inputs the dense branch retains full capacity. The same parameterization extends to low-rank adaptation by replacing $\M$ with $\tfrac{\alpha}{r}\B\A$.

\begin{figure}[t]
    \centering
    \includegraphics[width=\textwidth]{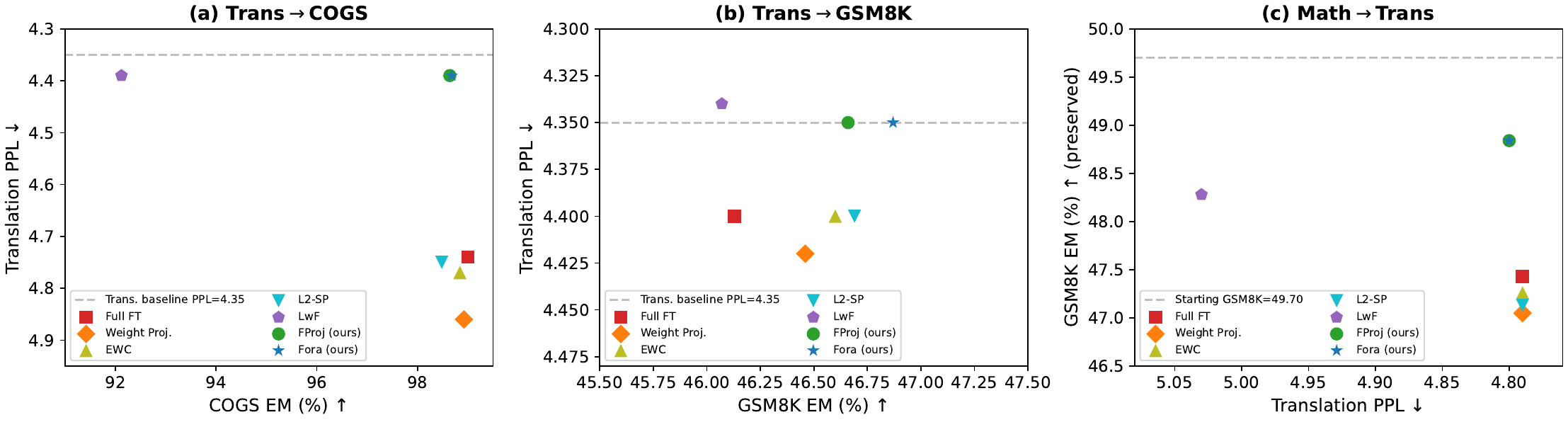}
    \caption{\textbf{Preservation--adaptation trade-off across all three settings} (visualizing Tables~\ref{tab:cogs_main}, \ref{tab:gsm8k_main}, and~\ref{tab:math_pres}). Each panel plots a new-task metric (horizontal) against the preserved-capability metric (vertical); the dashed line marks the preserved-capability target and the shaded band the favorable region. (a) Trans$\to$COGS and (b) Trans$\to$GSM8K plot new-task EM against translation PPL (lower better, axis inverted); (c) Math$\to$Trans plots translation PPL against retained GSM8K EM (higher better). In all three, \FProj/\Fora (large outlined markers) achieve the strongest preservation while remaining competitive on the new task, whereas the weight-space and regularization baselines sacrifice preservation.}
    \label{fig:tradeoff}
\end{figure}

\paragraph{Findings.} We instantiate the method on Qwen3-1.7B across three full-rank settings (Section~\ref{sec:experiments}). Learning COGS while preserving translation, we hold perplexity at 4.39---within 0.04 of the 4.35 baseline---at 98.6\% COGS exact match, while weight-space projection degrades perplexity to 4.86 at the same accuracy and is no better than unprotected fine-tuning. Learning GSM8K while preserving translation, we match the baseline perplexity exactly (4.35). Reversing the roles---learning translation while preserving math---we retain GSM8K within 0.9 points of the original model where full fine-tuning loses 2.3, confirming $\Q$ is not a translation-specific artifact. A controlled ablation swapping only $\PV$ for $\PQ$ shows the gain comes from \emph{what} is protected, not from projecting at all. Figure~\ref{fig:tradeoff} summarizes all three settings on the preservation--adaptation plane.

\paragraph{Contributions.}
(i) We reformulate capability preservation as \emph{function-space protection}: the protected object is the activation subspace induced by the retained capability, rather than parameter distance or the singular geometry of $\Wzero$.
(ii) We propose \Fora{} (\emph{Function-space Orthogonal Residual Adaptation}), a structurally constrained full-rank update
\[
\dW = \PU \M \PQ + \U_2 \D_\delta \V_2^\top ,
\]
where $\PQ$ blocks reads from capability-sensitive directions, $\PU$ blocks writes to dominant output directions, and the spectral channel provides bounded plasticity.
(iii) We characterize this parameterization through three operator-level properties and validate the mechanism with controlled $\PV$-versus-$\PQ$ ablations, showing that the key factor is the \emph{source} of the protected subspace. Section~\ref{sec:method} develops the method, Section~\ref{sec:experiments} evaluates the full-rank setting, and later sections present ablations and the low-rank extension.

%% file: sec/2_related_work.tex

\section{Related Work}
\label{sec:related}

\paragraph{Regularization and distillation.}
Classical continual-learning methods bound how far parameters or outputs may move. Importance-based regularizers such as EWC~\citep{kirkpatrick2017overcoming}, SI~\citep{zenke2017si}, and MAS~\citep{aljundi2018mas} estimate which parameters should remain stable; L2-SP~\citep{li2018explicit} penalizes the $\ell_2$ distance to the pre-trained weights; LwF~\citep{li2017learning} distills the frozen model's output distribution on new-task inputs. All three rely on \emph{indirect} signals---parameter importance, parameter proximity, output consistency---and none constrains which input directions an update may act on, leaving the capability's functional pathway exposed. Under full-rank fine-tuning they degrade translation almost as much as no protection at all (Section~\ref{sec:exp_cogs}).

\paragraph{Subspace and gradient projection.}
A second family constrains gradients. Gradient Projection Memory~\citep{saha2021gradient} projects new gradients onto the orthogonal complement of stored per-task subspaces; memory-based methods such as GEM~\citep{lopez2017gem} and A-GEM~\citep{chaudhry2019agem} constrain gradients using episodic replay buffers, providing strong baselines when preserved-capability data can be stored. O-LoRA~\citep{wang2023olora} learns sequential tasks in distinct orthogonal LoRA subspaces to minimize interference, and OPLoRA~\citep{wang2024oplora} provides the analogous projection in the PEFT setting. These act at the \emph{optimization} level, modifying gradients during training; we act at the \emph{parameterization} level, constraining the structure of $\dW$ so protection holds regardless of optimizer state or training trajectory.

\paragraph{Weight-space constrained fine-tuning.}
A third line decomposes $\Wzero$ by SVD and restricts adaptation to chosen spectral components: PiSSA~\citep{meng2024pissa} fine-tunes the residual after initializing from principal singular components, MiLoRA~\citep{wang2024milora} instead initializes from minor singular components, AdaLoRA~\citep{zhang2023adalora} adaptively allocates rank across layers by pruning singular values during training, and DoRA~\citep{liu2024dora} splits weights into magnitude and direction. Orthogonal fine-tuning methods such as BOFT~\citep{liu2024boft} parameterize update transformations to preserve orthogonality, but do not derive the protected subspace from a specific preserved capability's activations. Memory-efficient PEFT variants like QLoRA~\citep{dettmers2023qlora} focus on the computational cost of adaptation, whereas we address capability preservation under structural projection. All assume the weight's \emph{singular directions} define what to protect---but the top-$r$ right singular vectors $\V_1$ mark where the weight is large, not where a capability is active. Section~\ref{sec:exp_ablation} shows this mismatch directly: protecting $\V_1$ fails where the capability-derived $\Q$ succeeds.

\paragraph{Activation-aware adaptation.}
Recent adapters use activation statistics: CorDA~\citep{wang2024corda} orients low-rank initialization by input-activation covariance, and EVA~\citep{yang2024eva} guides placement by activation variance. At the predictive-function level, function-space regularization~\citep{rudner2023function} constrains the outputs a model may produce, while we constrain the \emph{internal} activation subspace through a hard projector $\PQ$ on per-layer weight updates. Mechanistic studies further suggest that specific behaviors such as refusal can be mediated by low-dimensional activation directions~\citep{arditi2024refusal}, supporting the premise that activation subspaces are meaningful objects of protection. We share the premise that activations encode capabilities but differ in two ways: we use the activation subspace as a \emph{hard right projector} $\PQ$ on the update operator rather than a soft initialization, and we establish the principle in \emph{full-rank} fine-tuning before specializing to low rank. To our knowledge this is the first controlled comparison of weight-space ($\PV$) against function-space ($\PQ$) projection under full-rank fine-tuning, identifying the projection \emph{source} as the decisive factor.

%% file: sec/3_method.tex

\section{Method}
\label{sec:method}

\begin{figure}[t]
    \centering
    \includegraphics[width=\textwidth]{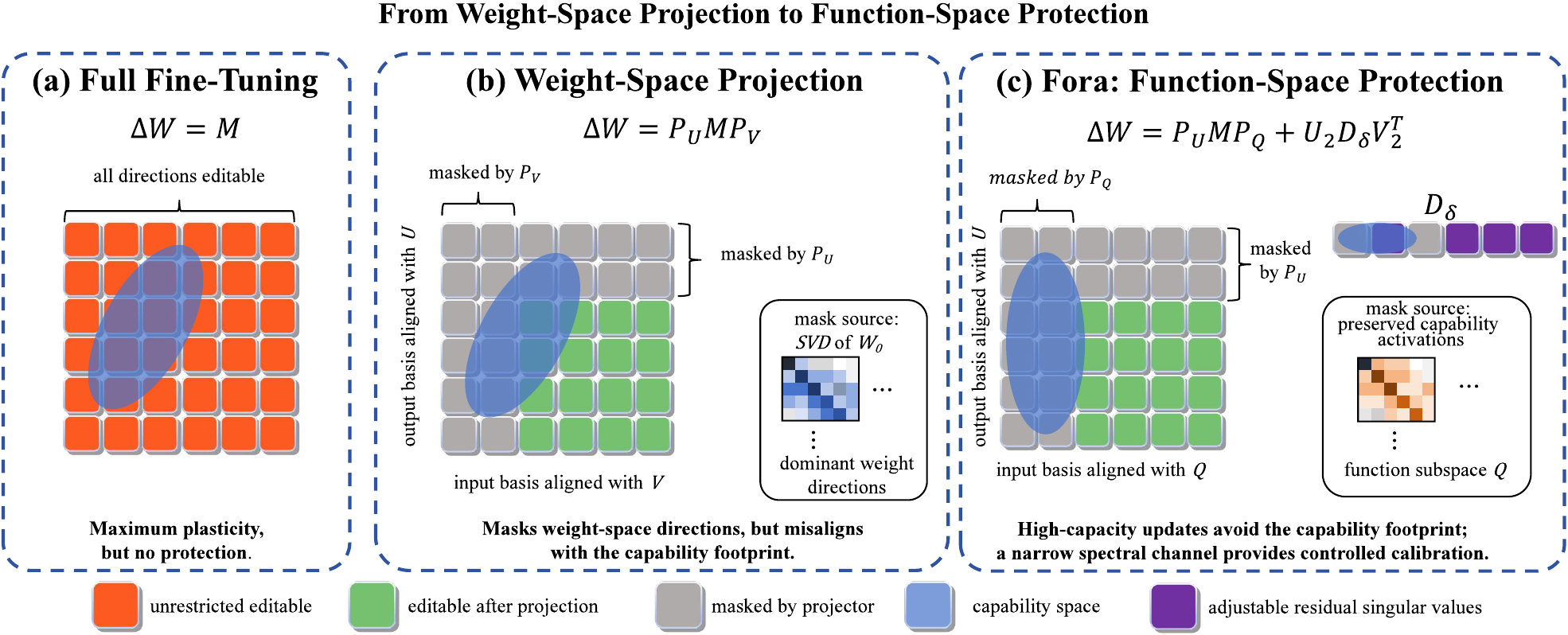}
    \caption{\textbf{From weight-space projection to function-space protection.}
    (a) Unconstrained full fine-tuning can overwrite any direction, offering maximal plasticity but no protection.
    (b) Weight-space projection masks the principal singular directions of $\Wzero$; these mark where the weight is large, which need not coincide with the capability's functional footprint.
    (c) \Fora{} replaces the weight-derived mask with a capability-derived projector $\PQ$, so the high-capacity branch avoids the activation directions the preserved capability uses, while a narrow spectral channel $\U_2\D_\delta\V_2^\top$ supplies controlled calibration.}
    \label{fig:overview}
\end{figure}

We develop the method as a sequence of parameterizations, each resolving a limitation of the preceding one (Figure~\ref{fig:overview}). The endpoint is a full-rank update that protects a capability through the activation subspace it induces, and the derivation shows that each design choice follows from the limitation it addresses rather than being assumed.

\subsection{Setup: the need for a constraint}
\label{sec:setup}

Consider a pre-trained weight $\Wzero \in \R^{d_{\text{out}} \times d_{\text{in}}}$ in any linear layer of a transformer. We are given a preserved capability $\Dpres$ (inputs only---no labels) and a new task $\Dnew$, and seek an update $\dW$ such that $\hW = \Wzero + \dW$ learns $\Dnew$ while retaining performance on $\Dpres$. Plain full fine-tuning sets
\begin{equation}
\dW = \M, \qquad \M \in \R^{d_{\text{out}} \times d_{\text{in}}},
\end{equation}
a dense matrix optimized on $\Dnew$. Its capacity is maximal, but so is its reach: $\M$ may read from and write to every input and output direction, including those on which the preserved capability depends. A constraint is therefore required, together with a principled basis from which to derive it.

\paragraph{SVD coordinates.}
A natural starting point is the structure already in $\Wzero$. Write $\Wzero = \U\Sigma\V^\top$ and split the singular triplets into a principal block (top-$r$) and a residual block:
\begin{equation}
\U = [\U_1\;\U_2],\quad \V = [\V_1\;\V_2],\quad \Sigma = \begin{bmatrix}\Sigma_1 & 0\\ 0 & \Sigma_2\end{bmatrix},
\end{equation}
with $\U_1 \in \R^{d_{\text{out}}\times r}$, $\V_1 \in \R^{d_{\text{in}}\times r}$. Define projectors onto the residual subspaces,
\begin{equation}
\PU = \I - \U_1\U_1^\top, \qquad \PV = \I - \V_1\V_1^\top .
\end{equation}
These give a vocabulary for restricting $\dW$ to the part of $\Wzero$ not dominated by its largest singular modes. It remains to choose the form this restriction should take.

\subsection{Spectral calibration and residual-space expansion}
\label{sec:spectral}

\paragraph{A conservative start.}
The most cautious update confined to the residual subspace is a diagonal rescaling of the residual singular values:
\begin{equation}\label{eq:delta_only}
\dW = \U_2\D_\delta\V_2^\top, \qquad \D_\delta = \diag(\bdelta),
\end{equation}
with $d_{\text{res}} = \min(d_{\text{out}},d_{\text{in}}) - r$ free scalars. The update lies entirely within $\U_2(\cdot)\V_2^\top$ and cannot affect the principal directions of $\Wzero$. It is safe, and---as we retain throughout---it constitutes a genuine, low-degree-of-freedom adjustment channel (Appendix~\ref{app:spectral_only} quantifies what this channel achieves in isolation). Its limitation is expressiveness: a diagonal map scales each residual mode independently and cannot represent interactions among them, so by itself it lacks the capacity to learn a demanding new task.

\paragraph{Adding mixing capacity.}
Relaxing the diagonal constraint restores expressiveness:
\begin{equation}\label{eq:res_expand}
\dW = \U_2(\D_\delta + \Sres)\V_2^\top,
\end{equation}
where $\Sres \in \R^{d_{\text{res}}\times d_{\text{res}}}$ mixes residual directions and $\D_\delta$ is retained as an explicit per-mode term. This form has genuine learning capacity while remaining within the residual subspace. However, $\Sres$ is expressed in the abstract SVD coordinates $(\U_2,\V_2)$, in which it is difficult to reason about \emph{which input directions} the update can access---precisely the reasoning required to protect a capability.

\subsection{From residual coordinates to function-space projection}
\label{sec:fproj}

\paragraph{Exposing the right projector.}
A standard identity rewrites the residual-coordinate form as a two-sided projection. For any $\M$,
\begin{equation}\label{eq:identity}
\U_2(\U_2^\top\M\V_2)\V_2^\top = (\I-\U_1\U_1^\top)\,\M\,(\I-\V_1\V_1^\top) = \PU\M\PV .
\end{equation}
Setting $\Sres = \U_2^\top\M\V_2$ turns Eq.~\ref{eq:res_expand} into
\begin{equation}\label{eq:bridge}
\U_2(\D_\delta+\Sres)\V_2^\top = \U_2\D_\delta\V_2^\top + \PU\M\PV .
\end{equation}
This does not change the set of representable updates---an unrestricted $\Sres$ and an unrestricted $\M$ span the same residual matrices---but it changes what is \emph{visible}: the term $\PU\M\PV$ displays the right projector $\PV$ explicitly. The constraint is no longer the opaque ``live in $\U_2(\cdot)\V_2^\top$'' but the functional ``do not read from $\calspan(\V_1)$.'' This exposes the question the SVD form concealed: whether $\calspan(\V_1)$ is in fact the right subspace to protect.

\paragraph{Limitation of $\PV$ as the protected subspace.}
$\V_1$ holds the input directions in which $\Wzero$ has its largest singular values---an aggregate compressed across every capability the model carries through a single factorization. A specific capability such as English--Chinese translation may instead activate patterns distributed over many moderate singular components, or oriented diagonally with respect to the SVD basis. $\V_1$ thus identifies where the \emph{weight} is large, not where the \emph{capability} is active.

\paragraph{Constructing a capability-specific projector.}
We therefore derive the protected subspace from the capability rather than the weight. Forwarding label-free calibration inputs from $\Dpres$, we collect the input activations at each target layer and stack them as $\X_{\text{pres}} \in \R^{N\times d}$ over $N$ calibration tokens. We form the activation covariance and take its top-$k$ eigenvectors,
\begin{equation}\label{eq:cov}
\C_{\text{pres}} = \tfrac{1}{N}\X_{\text{pres}}^\top\X_{\text{pres}}, \qquad \Q = [\bm{q}_1,\ldots,\bm{q}_k],
\end{equation}
ordered by decreasing eigenvalue---the directions of largest activation variance, i.e.\ the capability's functional signature in that layer. The \textbf{function-space right projector} is
\begin{equation}\label{eq:PQ}
\PQ = \I - \Q\Q^\top .
\end{equation}
Three properties are worth noting: the construction requires \emph{no labels} (inputs of $\Dpres$ suffice); it is performed \emph{per layer}; and it is \emph{capability-specific}, capturing the activation structure of the particular capability rather than an aggregate.

\begin{figure}[t]
    \centering
    \includegraphics[width=\textwidth]{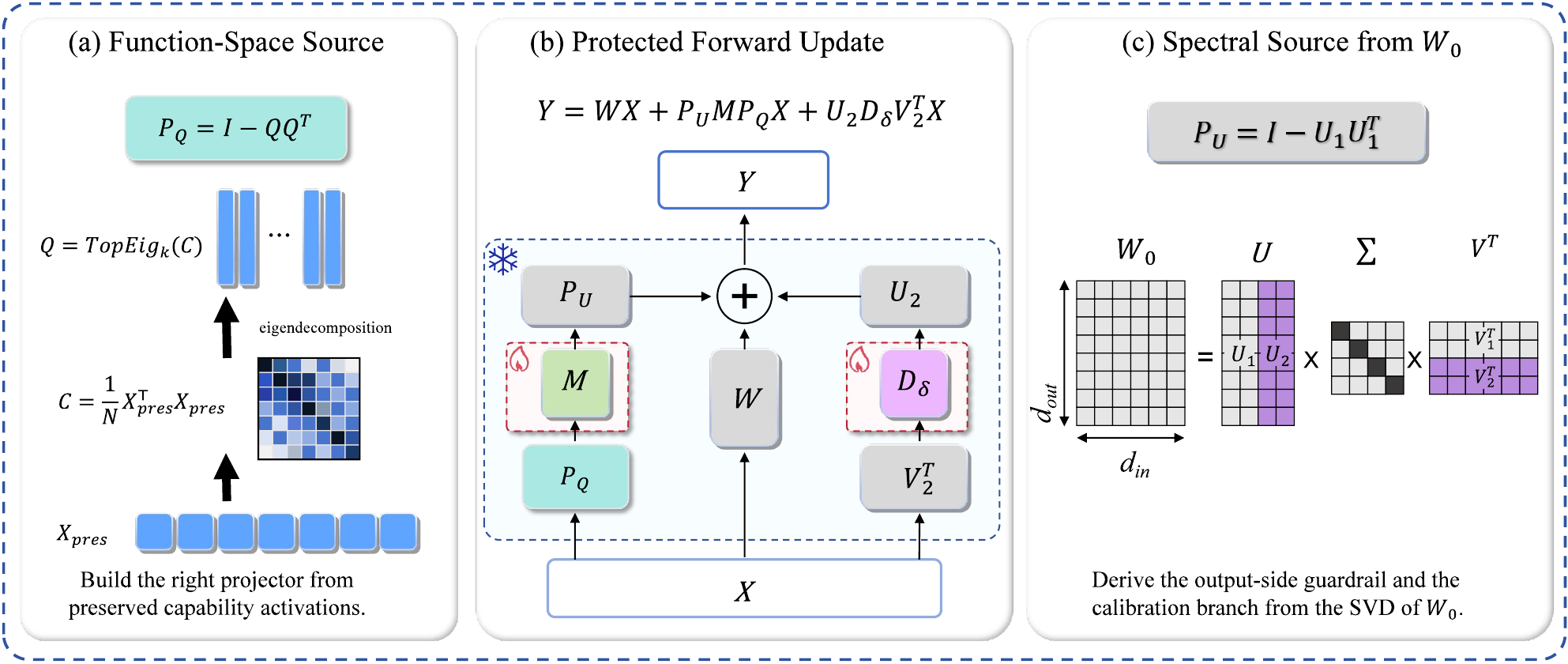}
    \caption{\textbf{Construction and forward computation of \Fora.}
    (a) \emph{Function-space source.} From label-free preserved-capability inputs we collect per-layer activations $\X_{\text{pres}}$, form the covariance $\C=\tfrac{1}{N}\X_{\text{pres}}^\top\X_{\text{pres}}$, and take its top-$k_f$ eigenvectors $\Q$ to build the right projector $\PQ=\I-\Q\Q^\top$.
    (b) \emph{Protected forward pass.} The output combines the frozen weight $\W$, the function-protected high-capacity branch $\PU\M\PQ$, and the spectral calibration branch $\U_2\D_\delta\V_2^\top$; only $\M$ and $\bdelta$ are trained.
    (c) \emph{Spectral source.} The left projector $\PU=\I-\U_1\U_1^\top$ and the calibration branch are both derived from the SVD of $\Wzero$, with $\U_1,\V_1$ the principal block and $\U_2,\V_2$ the residual block.}
    \label{fig:architecture}
\end{figure}

\paragraph{Function-space orthogonal residual adaptation (\Fora).}
Replacing $\PV$ by $\PQ$ in Eq.~\ref{eq:bridge} gives \Fora{} (\emph{Function-space Orthogonal Residual Adaptation}), our full-rank update:
\begin{equation}\label{eq:fpft}
\boxed{\;\dW_{\Fora} = \underbrace{\PU\,\M\,\PQ}_{\text{function-protected adaptation}} \;+\; \underbrace{\U_2\D_\delta\V_2^\top}_{\text{spectral calibration}}\;}
\end{equation}
$\PU\M\PQ$ is the high-capacity branch: a dense matrix with full expressive power, blocked from reading the capability's function directions (via $\PQ$) and from writing $\Wzero$'s principal output directions (via $\PU$). $\U_2\D_\delta\V_2^\top$ is the spectral channel from Eq.~\ref{eq:delta_only}. The projection-only variant without this channel, $\dW=\PU\M\PQ$, we denote \FProj. Table~\ref{tab:method_derivation} summarizes the derivation; Figure~\ref{fig:architecture} depicts the construction of $\Q$ and the resulting protected forward computation.

\paragraph{Structural properties.}
Three facts follow immediately from the projector definitions.

\begin{proposition}[Output-side protection]
\label{prop:output}
$\U_1^\top\,\PU\M\PQ = \bm{0}$: the high-capacity branch contributes nothing to the principal output directions of $\Wzero$.
\end{proposition}

\begin{proposition}[No read on capability directions]
\label{prop:zero}
For any $\bm{x}\in\calspan(\Q)$, $\;\PU\M\PQ\,\bm{x} = \bm{0}$: the high-capacity branch is blind to inputs aligned with the preserved capability's function directions.
\end{proposition}

\begin{proposition}[Controlled response via the spectral channel]
\label{prop:controlled}
Under the full update, for $\bm{x}\in\calspan(\Q)$,
\begin{equation}
\dW_{\Fora}\,\bm{x} = \U_2\D_\delta\V_2^\top\,\bm{x}.
\end{equation}
On protected directions the response is governed entirely by the spectral channel. The capability is not frozen: the model retains a low-degree-of-freedom, per-mode adjustment through $\bdelta$, while the dense branch is barred from these directions.
\end{proposition}

This statement makes the design intent precise: rather than hard-freezing the preserved capability, the method routes any adaptation along its directions through a narrow, controllable channel, while leaving the high-capacity branch unconstrained elsewhere.

\subsection{Two channels over complementary subspaces}
\label{sec:channels}

The final update decomposes the input space into two channels. The \textbf{function-protected channel} $\PU\M\PQ$ carries the majority of the learning capacity: between its two projectors the dense matrix $\M$ retains full expressive power for every direction outside $\calspan(\U_1)$ and $\calspan(\Q)$, with $\PU$ enforcing a weight-structure prior (preserving $\Wzero$'s dominant output space) and $\PQ$ enforcing the capability-specific constraint that distinguishes the method from weight-space projection. The \textbf{spectral channel} $\U_2\D_\delta\V_2^\top$ contributes only $d_{\text{res}}$ scalars, but by Proposition~\ref{prop:controlled} it is the only component able to respond to inputs in $\calspan(\Q)$, granting the protected capability a small, bounded degree of plasticity precisely where the dense branch is inactive, rather than freezing it. This yields a clear separation of roles: for $\bm{x}\in\calspan(\Q)$ only the spectral channel responds; for $\bm{x}\perp\calspan(\Q)$ the dense branch provides full capacity and the spectral channel adds residual calibration; and neither channel writes to $\calspan(\U_1)$. Protection is therefore a property of the parameterization rather than of the optimization trajectory.

\subsection{Low-rank extension}
\label{sec:lowrank}

The protection structure in Eq.~\ref{eq:fpft}---$\PU$ on the left, $\PQ$ on the right---does not depend on $\M$ being dense. Substituting a low-rank product,
\begin{equation}
\M \;\longrightarrow\; \tfrac{\alpha}{r}\B\A, \qquad \B\in\R^{d_{\text{out}}\times r_{\text{lora}}},\ \A\in\R^{r_{\text{lora}}\times d_{\text{in}}},\ r_{\text{lora}}\ll\min(d_{\text{out}},d_{\text{in}}),
\end{equation}
yields the parameter-efficient variant \Foralr{} (and \FProjlr{} without the spectral channel)
\begin{equation}\label{eq:fpft_lr}
\boxed{\;\dW_{\Foralr} = \PU\!\left(\tfrac{\alpha}{r}\B\A\right)\!\PQ \;+\; \U_2\D_\delta\V_2^\top\;}
\end{equation}
This is a substitution, not a new method: the principle is established in the full-rank setting and the low-rank form inherits it unchanged. We treat it as evidence of generality (Section~\ref{sec:exp_lowrank}), not the primary contribution.

\begin{table}[t]
\centering
\caption{Stages of the derivation. Each stage addresses a limitation of the previous one; the function-space update is the destination, with the low-rank form as a substitution.}
\label{tab:method_derivation}
\small
\begin{tabular}{@{}lll@{}}
\toprule
\textbf{Stage} & \textbf{Update rule} & \textbf{Limitation motivating the next stage} \\
\midrule
\FullFT                 & $\M$                                        & no protection \\
Spectral calibration    & $\U_2\D_\delta\V_2^\top$                    & diagonal: no mixing across residual modes \\
Residual expansion      & $\U_2(\D_\delta+\Sres)\V_2^\top$               & input/output structure hidden in SVD coords \\
Weight-space projection & $\U_2\D_\delta\V_2^\top + \PU\M\PV$        & $\PV$ protects weight geometry, not capability \\
\textbf{\Fora{} (ours)} & $\U_2\D_\delta\V_2^\top + \PU\M\PQ$ & --- \\
\Foralr{} (low-rank)   & $\U_2\D_\delta\V_2^\top + \PU\tfrac{\alpha}{r}\B\A\,\PQ$ & (parameter-efficient variant) \\
\bottomrule
\end{tabular}
\end{table}

%% file: sec/5_experiments.tex

\section{Experiments}
\label{sec:experiments}

We evaluate function-space protection across three full-rank settings, isolate the projection source by ablation, and verify the principle transfers to low-rank adaptation. The common question is whether protecting the \emph{capability-derived} subspace $\Q$ outperforms protecting the \emph{weight-derived} $\V_1$ while learning a new task; Figure~\ref{fig:tradeoff} previews the headline result, with our methods occupying the favorable corner of every preservation--adaptation plane.

\subsection{Setup}
\label{sec:exp_setup}

\paragraph{Model and protocol.}
All experiments use Qwen3-1.7B~\citep{yang2025qwen3} (decoder-only) in bf16, applying every method to all attention projections (Q, K, V, O) and feed-forward linear layers in each block. The function subspace $\Q$ is built per layer from the top $k_f = 16$ eigenvectors of the input-activation covariance, estimated from label-free preserved-capability inputs in one forward pass; the SVD rank for $\PU/\PV$ is $r = 100$. We optimize with AdamW~\citep{loshchilov2019decoupled} (lr $1\!\times\!10^{-5}$, batch 8, grad-accum 8) and early stopping (patience 5), selecting the best-validation-loss checkpoint. All full-rank methods train the same parameter count ($\sim\!1.7$B). Results are mean\,$\pm$\,std over 5 seeds unless noted. Full hyperparameters and dataset splits are given in Appendix~\ref{app:implementation} (Table~\ref{tab:datasets}), and per-seed results for all full-rank settings in Appendix~\ref{app:per_seed}.

\paragraph{Starting models.}
The starting weights $\Wzero$ are \emph{not} raw Qwen3-1.7B. To create a well-defined capability to preserve, we first LoRA-fine-tune the base model on 5{,}000 examples (translation or GSM8K) and merge the adapter into the weights; this capability-specialized checkpoint is $\Wzero$, and its merged capability is what we protect while learning the new task. The baseline preserved-capability numbers (translation chrF $=0.413$, PPL $=4.35$; GSM8K EM $=49.7$) are measured on these checkpoints.

\paragraph{Tasks and baselines.}
COGS~\citep{kim2020cogs} (compositional semantic parsing; 1000/300/500), GSM8K~\citep{cobbe2021gsm8k} (grade-school math; 1000/300/1319), and bidirectional English--Chinese translation (OPUS~\citep{tiedemann2012opus}; 1000/300/500), reporting COGS/GSM8K exact match (EM), translation chrF, and translation perplexity (PPL, lower better). Baselines: \FullFT ($\dW=\M$); \WProj ($\dW=\PU\M\PV$); \EWC~\citep{kirkpatrick2017overcoming}; \LtwoSP~\citep{li2018explicit}; \LwF~\citep{li2017learning}; and ours, \FProj ($\dW=\PU\M\PQ$) and \Fora ($\dW=\PU\M\PQ+\U_2\D_\delta\V_2^\top$).

\subsection{COGS Fine-Tuning with Translation Retention}
\label{sec:exp_cogs}

Our main setting starts from the translation-specialized $\Wzero$, builds $\Q$ from translation calibration inputs, and fine-tunes all linear layers on COGS. Table~\ref{tab:cogs_main} reports the result.

\begin{table}[t]
\centering\small
\caption{COGS Fine-Tuning with Translation Retention. All methods full-rank ($\sim\!1.7$B params), 5 seeds; the starting model is the translation-specialized $\Wzero$. \FProj{} ($\dW=\PU\M\PQ$) is the projection-only ablation; \Fora{} ($\dW=\PU\M\PQ+\U_2\D_\delta\V_2^\top$) is the full method.}
\label{tab:cogs_main}
\begin{tabular}{@{}llrrrrr@{}}
\toprule
Category & Method & COGS EM $\uparrow$ & chrF $\uparrow$ & $\Delta$chrF & PPL $\downarrow$ & Best Ep \\
\midrule
Baseline        & Translation        & ---                & $0.413_{\pm 0.010}$ & ---     & $4.35_{\pm 0.19}$ & --- \\
\midrule
No protection   & \FullFT  & $99.00_{\pm 0.45}$ & $0.405_{\pm 0.006}$ & $-0.01$ & $4.74_{\pm 0.24}$ & 6.0 \\
Weight-space    & \WProj     & $98.92_{\pm 0.58}$ & $0.405_{\pm 0.008}$ & $-0.01$ & $4.86_{\pm 0.22}$ & 4.4 \\
\midrule
Regularization  & \EWC       & $98.84_{\pm 0.61}$ & $0.404_{\pm 0.007}$ & $-0.01$ & $4.77_{\pm 0.21}$ & 6.0 \\
Regularization  & \LtwoSP      & $98.48_{\pm 0.86}$ & $0.405_{\pm 0.006}$ & $-0.01$ & $4.75_{\pm 0.21}$ & 5.0 \\
Regularization  & \LwF       & $92.12_{\pm 1.15}$ & $0.406_{\pm 0.007}$ & $-0.01$ & $4.39_{\pm 0.18}$ & 8.2 \\
\midrule
\multirow{2}{*}{Ours} & \FProj & $98.64_{\pm 0.99}$ & $0.414_{\pm 0.010}$ & $+0.00$ & $4.39_{\pm 0.19}$ & 5.0 \\
                               & \Fora  & $98.68_{\pm 0.81}$ & $0.414_{\pm 0.011}$ & $+0.00$ & $4.39_{\pm 0.19}$ & 4.8 \\
\bottomrule
\end{tabular}
\end{table}

We highlight three findings. \textbf{(1) $\PQ$ outperforms $\PV$.} Weight-space projection degrades translation PPL to 4.86---higher than even unprotected \FullFT (4.74)---whereas function-space projection maintains PPL at 4.39, within 0.04 of baseline, at 98.64\% COGS EM (only 0.36 below \FullFT). \textbf{(2) $\PQ$ outperforms regularization.} EWC and L2-SP reach PPL 4.75--4.77, statistically indistinguishable from unprotected fine-tuning. \textbf{(3) $\PQ$ avoids the distillation trade-off.} LwF attains PPL 4.39 but at a COGS EM of 92.12\% (versus $>98.5\%$ for all other methods), reflecting the stability--plasticity tension. Function-space protection is the only method that achieves both strong preservation and strong new-task learning.

\subsection{GSM8K Fine-Tuning with Translation Retention}
\label{sec:exp_gsm8k}

We repeat the setting with GSM8K as the new task. GSM8K is far harder than COGS for a 1.7B model trained on 1000 examples (best EM $\sim\!47\%$ vs $\sim\!99\%$), testing whether protection still helps when new-task headroom is narrow.

\begin{table}[t]
\centering\small
\caption{GSM8K Fine-Tuning with Translation Retention. Same translation-specialized $\Wzero$ as Table~\ref{tab:cogs_main} (chrF $0.413$, PPL $4.35$); the new task is GSM8K. 5 seeds.}
\label{tab:gsm8k_main}
\begin{tabular}{@{}lrrrrr@{}}
\toprule
Method & GSM8K EM $\uparrow$ & chrF $\uparrow$ & $\Delta$chrF & PPL $\downarrow$ & Best Ep \\
\midrule
Translation baseline & --- & $0.413_{\pm 0.010}$ & --- & $4.35_{\pm 0.19}$ & --- \\
\midrule
\FullFT & $46.13_{\pm 1.08}$ & $0.406_{\pm 0.010}$ & $-0.01$ & $4.40_{\pm 0.17}$ & 1.8 \\
\WProj    & $46.46_{\pm 0.96}$ & $0.406_{\pm 0.010}$ & $-0.01$ & $4.42_{\pm 0.18}$ & 1.8 \\
\EWC      & $46.60_{\pm 1.18}$ & $0.405_{\pm 0.010}$ & $-0.01$ & $4.40_{\pm 0.17}$ & 1.8 \\
\LtwoSP     & $46.69_{\pm 0.60}$ & $0.405_{\pm 0.010}$ & $-0.01$ & $4.40_{\pm 0.17}$ & 1.8 \\
\LwF      & $46.07_{\pm 0.78}$ & $0.407_{\pm 0.010}$ & $-0.01$ & $4.34_{\pm 0.17}$ & 2.8 \\
\midrule
\FProj & $46.66_{\pm 0.91}$ & $0.411_{\pm 0.010}$ & $-0.00$ & $4.35_{\pm 0.17}$ & 1.8 \\
\Fora  & $46.87_{\pm 1.01}$ & $0.411_{\pm 0.010}$ & $-0.00$ & $4.35_{\pm 0.17}$ & 1.8 \\
\bottomrule
\end{tabular}
\end{table}

GSM8K EM spans only 0.8 points across all methods (46.07--46.87), reflecting the task's difficulty at this scale; the discriminating axis is therefore preservation. \FProj and \Fora maintain chrF at 0.411 and PPL at 4.35---indistinguishable from baseline---while \FullFT, \WProj, \EWC, and \LtwoSP all drift (chrF 0.405--0.406, PPL 4.40--4.42). \Fora additionally attains the highest GSM8K EM (46.87): when new-task headroom is limited, the spectral channel's modest additional plasticity yields a small improvement without disturbing preservation. As Figure~\ref{fig:tradeoff}(b) shows, our methods occupy the favorable region in this setting as well.

\subsection{Translation Fine-Tuning with Math Reasoning Retention}
\label{sec:exp_math}

To show that $\Q$ is not specific to translation, we reverse the roles: beginning from a GSM8K-fine-tuned model, we learn translation as the new task while preserving math reasoning, with $\Q=\Q_{\text{math}}$ constructed from GSM8K calibration inputs. This is the more demanding direction and the one most likely to reveal forgetting.

\begin{table}[t]
\centering\small
\caption{Translation Fine-Tuning with Math Reasoning Retention. Starting model: GSM8K-tuned Qwen3-1.7B; $\Q=\Q_{\text{math}}$ from GSM8K inputs; 5 seeds. $\Delta$GSM8K is measured against the GSM8K-tuned starting model (49.7).}
\label{tab:math_pres}
\begin{tabular}{@{}lrrrrr@{}}
\toprule
Method & chrF $\uparrow$ & PPL $\downarrow$ & GSM8K EM $\uparrow$ & $\Delta$GSM8K & Best Ep \\
\midrule
Starting model & --- & --- & 49.70 & --- & --- \\
\midrule
\FullFT & $0.396_{\pm 0.008}$ & $4.79_{\pm 0.22}$ & $47.43_{\pm 0.68}$ & $-2.27$ & 1.0 \\
\WProj    & $0.395_{\pm 0.008}$ & $4.79_{\pm 0.22}$ & $47.05_{\pm 0.84}$ & $-2.65$ & 1.0 \\
\EWC      & $0.395_{\pm 0.009}$ & $4.79_{\pm 0.22}$ & $47.26_{\pm 0.80}$ & $-2.44$ & 1.0 \\
\LtwoSP     & $0.394_{\pm 0.007}$ & $4.79_{\pm 0.22}$ & $47.13_{\pm 0.47}$ & $-2.57$ & 1.0 \\
\LwF      & $0.395_{\pm 0.007}$ & $5.03_{\pm 0.22}$ & $48.28_{\pm 0.68}$ & $-1.42$ & 2.2 \\
\midrule
\FProj & $0.390_{\pm 0.008}$ & $4.80_{\pm 0.22}$ & $48.84_{\pm 0.36}$ & $-0.86$ & 1.0 \\
\Fora  & $0.390_{\pm 0.009}$ & $4.80_{\pm 0.22}$ & $48.84_{\pm 0.42}$ & $-0.86$ & 1.0 \\
\bottomrule
\end{tabular}
\end{table}

Table~\ref{tab:math_pres} confirms that the pattern extends to math. Our methods retain GSM8K at 48.84\% ($\Delta=-0.86$), versus $-2.27$ for \FullFT and $-2.4$ to $-2.7$ for the weight-space and regularization baselines---approximately a threefold reduction in forgetting. LwF is intermediate ($\Delta=-1.42$) but degrades translation (PPL 5.03 vs.\ 4.79--4.80). The sole trade-off is a small reduction in chrF for \FProj/\Fora (0.390 vs.\ 0.396): when the preserved and new capabilities share activation structure, protecting $\Q_{\text{math}}$ withholds a portion of new-task capacity, a modest and deliberate cost given the substantial retention gain. Importantly, constructing $\Q$ from math and preserving math excludes the alternative explanation that $\Q$ merely encodes translation.

\subsection{Ablation: The Projection Source Is What Matters}
\label{sec:exp_ablation}

\paragraph{$\PV$ versus $\PQ$.}
The central mechanistic claim is that projecting is not enough---\emph{what} subspace is projected onto decides preservation. Table~\ref{tab:proj_source} isolates this by swapping only the right projector ($\PV\!\to\!\PQ$) while holding $\PU$ and the high-capacity branch $\M$ fixed, across all three settings.

\begin{table}[t]
\centering\small
\caption{Projection Source Ablation: Weight-Space vs.\ Function-Space. Swapping only the right projector $\PV\!\to\!\PQ$ while holding $\PU$ and $\M$ fixed. $\PQ$ wins on the preserved metric in every setting.}
\label{tab:proj_source}
\begin{tabular}{@{}llrr@{}}
\toprule
Setting & Right projector & New task $\uparrow$ & Preserved \\
\midrule
\multirow{2}{*}{Trans$\to$COGS}  & $\PV$ (weight)   & COGS EM $98.92$ & PPL $4.86$ \\
                                 & $\PQ$ (function) & COGS EM $98.64$ & PPL $4.39$ \\
\midrule
\multirow{2}{*}{Trans$\to$GSM8K} & $\PV$ (weight)   & GSM8K EM $46.46$ & PPL $4.42$ \\
                                 & $\PQ$ (function) & GSM8K EM $46.66$ & PPL $4.35$ \\
\midrule
\multirow{2}{*}{Math$\to$Trans}  & $\PV$ (weight)   & chrF $0.395$ & GSM8K EM $47.05$ \\
                                 & $\PQ$ (function) & chrF $0.390$ & GSM8K EM $48.84$ \\
\bottomrule
\end{tabular}
\end{table}

In every setting $\PQ$ improves the preserved metric over $\PV$: PPL decreases by 0.47 (COGS) and 0.07 (GSM8K), and GSM8K retention increases by 1.79 points (Math$\to$Trans); new-task performance is comparable or marginally lower, the expected consequence of protecting capability-relevant directions more precisely (Figure~\ref{fig:proj_source}, Appendix~\ref{app:peft_full}). We conclude that \textbf{projection alone is insufficient; the projection \emph{source}---weight geometry versus capability function---is the decisive factor.}

\paragraph{Contribution of the spectral channel.}
Table~\ref{tab:delta_ablation} contrasts \FProj (without $\bdelta$) and \Fora (with $\bdelta$). Adding the channel never degrades a preserved metric, and where the new task admits additional plasticity it produces a small, consistent improvement---most evident as $+0.21$ GSM8K EM in the headroom-limited Trans$\to$GSM8K setting and $+0.04$ on COGS. This matches the behavior characterized by Proposition~\ref{prop:controlled}: a bounded calibration channel that the optimizer can exploit when beneficial and that otherwise remains inactive, leaving the protection guarantee intact. We therefore retain $\bdelta$ as a low-cost component of the full method.

\begin{table}[t]
\centering\small
\caption{Spectral Channel Ablation: \FProj{} vs.\ \Fora. \FProj: $\dW=\PU\M\PQ$; \Fora: $\dW=\PU\M\PQ+\U_2\D_\delta\V_2^\top$. The channel adds a small new-task margin where headroom exists and never harms preservation.}
\label{tab:delta_ablation}
\begin{tabular}{@{}lrrr@{}}
\toprule
Setting / metric & \FProj & \Fora & $\Delta$ \\
\midrule
COGS EM                  & $98.64$ & $98.68$ & $+0.04$ \\
COGS translation PPL     & $4.39$  & $4.39$  & $\;\,0.00$ \\
GSM8K EM                 & $46.66$ & $46.87$ & $+0.21$ \\
GSM8K translation PPL    & $4.35$  & $4.35$  & $\;\,0.00$ \\
Math$\to$Trans chrF      & $0.3897$ & $0.3904$ & $+0.0007$ \\
Math$\to$Trans GSM8K EM  & $48.84$ & $48.84$ & $\;\,0.00$ \\
\bottomrule
\end{tabular}
\end{table}

\paragraph{Subspace dimensions $r$ and $k_f$.}
The method has two structural dimensions: the SVD frozen rank $r$ in $\PU$ (default 100) and the function-subspace size $k_f$ in $\PQ$ (default 16). Sweeping each independently on the Math$\to$Trans setting (5 seeds per configuration), GSM8K retention stays within a $0.4$-point band as $r$ ranges over $20$--$400$ and $k_f$ over $4$--$64$, with translation chrF/PPL unchanged---every configuration lies within one standard deviation of the default, so $(r,k_f)=(100,16)$ is not a tuned operating point. This robustness concerns \emph{how many} directions are protected, not \emph{which} subspace they come from; the latter remains decisive (Table~\ref{tab:proj_source}). Full results are in Appendix~\ref{app:dim_sweep} (Table~\ref{tab:dim_sweep}).

\subsection{Low-Rank Extension}
\label{sec:exp_lowrank}

Finally, we verify that the principle transfers to parameter-efficient fine-tuning. Table~\ref{tab:lowrank} compares methods at matched parameter count ($\sim\!17$M) under COGS fine-tuning with translation retention.

\begin{table}[t]
\centering\small
\caption{Low-Rank Extension under COGS Fine-Tuning. Replacing the full-rank matrix $\M$ with a LoRA~\citep{hu2021lora} low-rank product $\tfrac{\alpha}{r}\B\A$; all methods $\sim\!17$M trainable parameters, 5 seeds. \FProjlr: $\dW=\PU\tfrac{\alpha}{r}\B\A\,\PQ$; \Foralr{} adds the spectral channel $\U_2\D_\delta\V_2^\top$. This table evaluates generality, not the main full-rank claim. Baseline PPL 4.35.}
\label{tab:lowrank}
\begin{tabular}{@{}lrrr@{}}
\toprule
Method & COGS EM $\uparrow$ & chrF $\uparrow$ & PPL $\downarrow$ \\
\midrule
LoRA-r16             & $95.44_{\pm 1.75}$ & $0.414_{\pm 0.010}$ & $4.54_{\pm 0.20}$ \\
LoRA-r16 + replay 1\% & $95.52_{\pm 1.78}$ & $0.414_{\pm 0.009}$ & $4.50_{\pm 0.19}$ \\
LoRA-r16 + replay 10\% & $95.84_{\pm 1.77}$ & $0.405_{\pm 0.010}$ & $4.55_{\pm 0.17}$ \\
OPLoRA-r16           & $94.44_{\pm 2.55}$ & $0.414_{\pm 0.010}$ & $4.54_{\pm 0.18}$ \\
CorDA-k16-r16        & $94.16_{\pm 1.40}$ & $0.414_{\pm 0.009}$ & $4.38_{\pm 0.19}$ \\
\midrule
\FProjlr         & $94.60_{\pm 2.29}$ & $0.413_{\pm 0.010}$ & $4.38_{\pm 0.18}$ \\
\Foralr & $94.48_{\pm 2.21}$ & $0.413_{\pm 0.009}$ & $4.38_{\pm 0.19}$ \\
\bottomrule
\end{tabular}
\end{table}

The protection mechanism behaves identically at low rank. \Foralr{} and its projection-only variant \FProjlr{} match CorDA on translation PPL (4.38, closest to baseline) and exceed OPLoRA's weight-space projection on COGS EM (94.60 vs.\ 94.44) at lower PPL. Replay-based LoRA attains the highest COGS EM (95.84\% at 10\% replay), but at higher PPL (4.50--4.55) and with a requirement for preserved-capability labels and mixed training that function-space protection does not impose. We present this not as a contribution to PEFT but as evidence that the same $\PU(\cdot)\PQ$ structure preserves capability whether the adapter is dense or low-rank; the full-rank results (Tables~\ref{tab:cogs_main}--\ref{tab:math_pres}) remain the primary contribution.

%% file: sec/6_discussion.tex

\section{Discussion}
\label{sec:discussion}

\paragraph{Why weight-space projection fails.}
The SVD of $\Wzero$ captures the weight's dominant linear modes---an aggregate over every capability the model holds---so a single capability may reside along moderate-singular-value directions that the top-$r$ block excludes. Removing $\calspan(\V_1)$ then has two adverse effects: it leaves capability-critical directions outside the block unprotected, and it constrains in-block directions that the capability does not use. $\PQ$ avoids both by acting on the input space the capability \emph{actually activates}, which accounts for its retaining the preserved capability across all three settings where $\PV$ does not.

\paragraph{Two complementary mechanisms.}
Protection and plasticity are assigned to separate components: the projectors $(\PU,\PQ)$ provide structural protection, while the spectral channel $\bdelta$---the only component permitted to act on the protected directions (Proposition~\ref{prop:controlled})---provides a bounded degree of plasticity there. Empirically, this channel yields a small, consistent new-task improvement where the task admits headroom and remains inactive otherwise, never reducing preservation (Table~\ref{tab:delta_ablation}); Appendix~\ref{app:spectral_only} confirms that it carries real but limited plasticity in isolation. The two roles are thus complementary rather than competing, so the method gains capacity without compromising the protection guarantee.

\paragraph{Scope and generality.}
Full-rank methods reach 98.5--99.0\% COGS EM, against 95.8\% for the best low-rank method (Tables~\ref{tab:cogs_main},~\ref{tab:lowrank})---a $\sim\!3$-point plasticity gap under an identical protection mechanism, leading us to recommend full-rank fine-tuning where resources permit and the low-rank form otherwise. Finally, $\Q$ is not specific to translation: constructed from GSM8K inputs, $\Q_{\text{math}}$ preserves math reasoning during translation fine-tuning (Table~\ref{tab:math_pres}). As the construction requires only label-free inputs, function-space protection applies to any preservation target for which representative calibration data are available.

%% file: sec/7_limitations.tex

\section{Limitations and Future Work}
\label{sec:limitations}

\textbf{Single-capability protection.} The current formulation protects one capability through a single $\Q$; preserving several simultaneously requires a union projector $\PQ = \I - [\Q_1\;\Q_2][\Q_1\;\Q_2]^\top$, which raises the effective dimension and may reduce the high-capacity branch's adaptive room. \textbf{Subspace dimensions.} \Fora{} is robust to both structural dimensions on the setting we sweep (Table~\ref{tab:dim_sweep}); whether this insensitivity holds across tasks, and whether an automatic eigenvalue-decay criterion can select $k_f$, remains open. \textbf{Calibration data.} $\Q$ depends on the calibration inputs; its sensitivity to their number (we use 500--1000) and to domain shift is not yet characterized. \textbf{Replay.} Replay remains strong at low rank (Table~\ref{tab:lowrank}); whether data-level and structural protection are complementary is a natural next step. \textbf{Scale and architecture.} All experiments use Qwen3-1.7B; larger models, other architectures, and additional capabilities would strengthen the generality claims. None of these issues bears on the central claim---that activation-derived function subspaces protect capabilities more faithfully than weight-space projections---but each identifies a direction for further study.

%% file: sec/8_conclusion.tex

\section{Conclusion}
\label{sec:conclusion}

We have argued that capability preservation in fine-tuning is more appropriately formulated as \textbf{function-space protection} than as weight-space protection. The activation-derived subspace $\Q$---constructed from preserved-capability inputs without labels---is a more faithful protection target than the weight-derived $\V_1$, as it captures the directions a capability uses rather than those in which the weight is large. Instantiated as $\dW = \PU\M\PQ + \U_2\D_\delta\V_2^\top$, the method preserves capabilities more effectively than weight-space projection and standard regularization across three settings, at the cost of only a small new-task trade-off in the math-preservation setting, and a controlled ablation attributes the improvement to the projection \emph{source} rather than to projection itself. Established in full-rank fine-tuning, the principle transfers without modification to low-rank adaptation via $\M\to\tfrac{\alpha}{r}\B\A$.

%% file: appendix/A_implementation.tex

\section{Implementation Details}
\label{app:implementation}

\subsection{Model Configuration}

\begin{itemize}[leftmargin=*,nosep]
\item \textbf{Base model}: Qwen3-1.7B~\citep{yang2025qwen3} (decoder-only transformer)
\item \textbf{Starting checkpoints} $\Wzero$: capability-specialized models obtained by LoRA fine-tuning the base model on 5{,}000 examples (translation or GSM8K) and merging the adapter into the weights. The merged capability is the one preserved during subsequent new-task fine-tuning.
\item \textbf{Precision}: bf16 (mixed-precision training and inference)
\item \textbf{Target modules}: All linear layers in attention (Q, K, V, O) and feed-forward networks (gate, up, down projections) across all 28 transformer blocks
\item \textbf{Frozen rank}: $r = 100$ for the SVD-based left projector $\PU$ (and $\PV$ in the weight-space baseline)
\item \textbf{Function subspace dimension}: $k_f = 16$ for $\Q$
\item \textbf{Calibration samples}: 500--1000 preserved-capability inputs (no labels)
\end{itemize}

\subsection{Training Configuration}

\begin{itemize}[leftmargin=*,nosep]
\item \textbf{Optimizer}: AdamW~\citep{loshchilov2019decoupled}
\item \textbf{Learning rate}: 1e-5 (full-rank), 1e-4 (low-rank)
\item \textbf{Batch size}: 8 per GPU
\item \textbf{Gradient accumulation}: 8 (effective batch size = 64)
\item \textbf{Max epochs}: 30 (full-rank); early stopping patience = 5
\item \textbf{LoRA configuration}: $r = 16$, $\alpha = 128$; target modules = all linear
\item \textbf{Seeds}: 42, 101, 102, 123, 456 (5 seeds for most experiments)
\end{itemize}

\subsection{Dataset Splits}

\begin{table}[htbp]
\centering
\caption{Dataset splits used in all experiments.}
\label{tab:datasets}
\begin{tabular}{@{}lrrr@{}}
\toprule
\textbf{Dataset} & \textbf{Train} & \textbf{Valid} & \textbf{Test} \\
\midrule
COGS         & 1000 & 300 & 500 \\
GSM8K        & 1000 & 300 & 1319 (full official) \\
Translation  & 1000 & 300 & 500 (bidirectional) \\
\bottomrule
\end{tabular}
\end{table}

\subsection{Evaluation Protocol}

\begin{itemize}[leftmargin=*,nosep]
\item \textbf{Checkpoint selection}: Best validation loss on the new task.
\item \textbf{COGS EM}: Exact string match between generated and reference output, evaluated on 500 test examples.
\item \textbf{GSM8K EM}: Exact match on final answer number, evaluated on 1319 test examples with greedy decoding.
\item \textbf{Translation chrF/PPL}: chrF score and perplexity on 500 bidirectional test examples; PPL computed on response tokens only.
\item \textbf{Per-seed results}: Reported as mean $\pm$ standard deviation across seeds.
\end{itemize}

%% file: appendix/B_per_seed.tex

\section{Full Per-Seed Results}
\label{app:per_seed}

This appendix reports the individual-seed measurements underlying the aggregated means and standard deviations in Section~\ref{sec:experiments}: Table~\ref{tab:cogs_per_seed} for COGS fine-tuning (summarized in Table~\ref{tab:cogs_main}), Table~\ref{tab:gsm8k_per_seed} for GSM8K fine-tuning (Table~\ref{tab:gsm8k_main}), and Table~\ref{tab:math_per_seed} for the reverse translation-with-math-preservation direction (Table~\ref{tab:math_pres}).

\subsection{COGS Full Fine-Tuning (5 Seeds)}

\begin{table}[htbp]
\centering
\caption{Per-seed results for COGS full fine-tuning with translation preservation.}
\label{tab:cogs_per_seed}
\small
\begin{tabular}{@{}llrrrrr@{}}
\toprule
\textbf{Method} & \textbf{Seed} & \textbf{COGS EM} & \textbf{chrF} & \textbf{PPL} & \textbf{Val Loss} & \textbf{Best Ep} \\
\midrule
\multirow{5}{*}{\FullFT} & 42  & 98.80 & 0.4026 & 4.600 & 0.0025 & 6 \\
& 101 & 98.40 & 0.4141 & 4.505 & 0.0032 & 8 \\
& 102 & 99.80 & 0.4086 & 4.864 & 0.0025 & 6 \\
& 123 & 99.00 & 0.4017 & 5.029 & 0.0019 & 5 \\
& 456 & 99.00 & 0.3967 & 4.714 & 0.0025 & 5 \\
\midrule
\multirow{5}{*}{\WProj} & 42  & 99.20 & 0.4013 & 4.617 & 0.0029 & 5 \\
& 101 & 99.40 & 0.4138 & 4.660 & 0.0037 & 5 \\
& 102 & 98.40 & 0.4135 & 5.243 & 0.0022 & 3 \\
& 123 & 99.00 & 0.3961 & 4.954 & 0.0021 & 5 \\
& 456 & 98.60 & 0.4010 & 4.828 & 0.0030 & 4 \\
\midrule
\multirow{5}{*}{\FProj} & 42  & 98.60 & 0.4124 & 4.219 & 0.0024 & 5 \\
& 101 & 98.80 & 0.4242 & 4.153 & 0.0031 & 6 \\
& 102 & 97.40 & 0.4218 & 4.721 & 0.0022 & 4 \\
& 123 & 99.40 & 0.4017 & 4.478 & 0.0019 & 5 \\
& 456 & 99.00 & 0.4101 & 4.394 & 0.0021 & 5 \\
\midrule
\multirow{5}{*}{\Fora} & 42  & 98.60 & 0.4124 & 4.210 & 0.0024 & 5 \\
& 101 & 98.00 & 0.4271 & 4.154 & 0.0030 & 5 \\
& 102 & 98.00 & 0.4219 & 4.721 & 0.0024 & 4 \\
& 123 & 99.40 & 0.4010 & 4.478 & 0.0019 & 5 \\
& 456 & 99.40 & 0.4098 & 4.386 & 0.0021 & 5 \\
\bottomrule
\end{tabular}
\end{table}

\subsection{GSM8K Full Fine-Tuning (5 Seeds)}

\begin{table}[htbp]
\centering
\caption{Per-seed results for GSM8K full fine-tuning with translation preservation.}
\label{tab:gsm8k_per_seed}
\small
\begin{tabular}{@{}llrrrrr@{}}
\toprule
\textbf{Method} & \textbf{Seed} & \textbf{GSM8K EM} & \textbf{chrF} & \textbf{PPL} & \textbf{Val Loss} & \textbf{Best Ep} \\
\midrule
\multirow{5}{*}{\FullFT} & 101 & 47.54 & 0.4079 & 4.270 & 0.432 & 1 \\
& 102 & 47.23 & 0.4102 & 4.220 & 0.433 & 2 \\
& 123 & 45.72 & 0.4179 & 4.291 & 0.445 & 2 \\
& 42  & 45.41 & 0.3994 & 4.568 & 0.422 & 2 \\
& 456 & 44.73 & 0.3928 & 4.647 & 0.447 & 2 \\
\midrule
\multirow{5}{*}{\FProj} & 101 & 47.23 & 0.4173 & 4.231 & 0.433 & 1 \\
& 102 & 47.23 & 0.4150 & 4.174 & 0.433 & 2 \\
& 123 & 45.64 & 0.4189 & 4.233 & 0.447 & 2 \\
& 42  & 47.69 & 0.4030 & 4.502 & 0.421 & 2 \\
& 456 & 45.49 & 0.4000 & 4.598 & 0.449 & 2 \\
\midrule
\multirow{5}{*}{\Fora} & 101 & 47.23 & 0.4178 & 4.235 & 0.433 & 1 \\
& 102 & 44.96 & 0.4152 & 4.175 & 0.435 & 2 \\
& 123 & 47.92 & 0.4200 & 4.233 & 0.443 & 2 \\
& 42  & 46.85 & 0.4029 & 4.501 & 0.421 & 2 \\
& 456 & 47.38 & 0.3988 & 4.598 & 0.448 & 2 \\
\bottomrule
\end{tabular}
\end{table}

\subsection{Translation Fine-Tuning with Math Preservation (5 Seeds)}

Table~\ref{tab:math_per_seed} reports the reverse direction: a GSM8K-tuned starting model, translation as the new task, and math reasoning preserved through $\Q_{\text{math}}$.

\begin{table}[htbp]
\centering
\caption{Per-seed results for translation fine-tuning with math-reasoning preservation. GSM8K is the preserved capability; chrF/PPL are the new translation task.}
\label{tab:math_per_seed}
\small
\begin{tabular}{@{}llrrrrr@{}}
\toprule
\textbf{Method} & \textbf{Seed} & \textbf{GSM8K EM} & \textbf{chrF} & \textbf{PPL} & \textbf{Val Loss} & \textbf{Best Ep} \\
\midrule
\multirow{5}{*}{\FullFT} & 101 & 46.55 & 0.4035 & 4.62 & 1.705 & 1 \\
& 102 & 47.01 & 0.3996 & 4.57 & 1.803 & 1 \\
& 123 & 48.14 & 0.4000 & 4.71 & 1.814 & 1 \\
& 42  & 48.07 & 0.3855 & 4.96 & 1.696 & 1 \\
& 456 & 47.38 & 0.3905 & 5.07 & 1.772 & 1 \\
\midrule
\multirow{5}{*}{\WProj} & 101 & 45.72 & 0.4010 & 4.62 & 1.705 & 1 \\
& 102 & 46.85 & 0.4008 & 4.57 & 1.803 & 1 \\
& 123 & 47.16 & 0.3983 & 4.71 & 1.813 & 1 \\
& 42  & 47.69 & 0.3831 & 4.96 & 1.696 & 1 \\
& 456 & 47.84 & 0.3909 & 5.08 & 1.772 & 1 \\
\midrule
\multirow{5}{*}{\EWC} & 101 & 46.25 & 0.4021 & 4.62 & 1.704 & 1 \\
& 102 & 46.93 & 0.3992 & 4.57 & 1.804 & 1 \\
& 123 & 47.31 & 0.4009 & 4.71 & 1.815 & 1 \\
& 42  & 48.45 & 0.3832 & 4.96 & 1.697 & 1 \\
& 456 & 47.38 & 0.3872 & 5.07 & 1.773 & 1 \\
\midrule
\multirow{5}{*}{\LtwoSP} & 101 & 46.55 & 0.4002 & 4.62 & 1.705 & 1 \\
& 102 & 46.93 & 0.3977 & 4.57 & 1.804 & 1 \\
& 123 & 47.23 & 0.3992 & 4.71 & 1.814 & 1 \\
& 42  & 47.84 & 0.3826 & 4.96 & 1.697 & 1 \\
& 456 & 47.08 & 0.3914 & 5.07 & 1.773 & 1 \\
\midrule
\multirow{5}{*}{\LwF} & 101 & 47.69 & 0.4003 & 4.86 & 1.772 & 3 \\
& 102 & 48.14 & 0.3998 & 4.80 & 1.866 & 2 \\
& 123 & 47.61 & 0.3976 & 4.97 & 1.879 & 2 \\
& 42  & 48.82 & 0.3900 & 5.21 & 1.751 & 2 \\
& 456 & 49.13 & 0.3854 & 5.32 & 1.829 & 2 \\
\midrule
\multirow{5}{*}{\FProj} & 101 & 48.60 & 0.3981 & 4.65 & 1.712 & 1 \\
& 102 & 48.75 & 0.3961 & 4.58 & 1.811 & 1 \\
& 123 & 49.36 & 0.3926 & 4.73 & 1.820 & 1 \\
& 42  & 48.45 & 0.3794 & 4.96 & 1.703 & 1 \\
& 456 & 49.05 & 0.3822 & 5.09 & 1.781 & 1 \\
\midrule
\multirow{5}{*}{\Fora} & 101 & 48.22 & 0.3976 & 4.65 & 1.711 & 1 \\
& 102 & 49.13 & 0.3956 & 4.58 & 1.811 & 1 \\
& 123 & 49.28 & 0.3965 & 4.73 & 1.821 & 1 \\
& 42  & 48.90 & 0.3784 & 4.96 & 1.703 & 1 \\
& 456 & 48.67 & 0.3841 & 5.09 & 1.782 & 1 \\
\bottomrule
\end{tabular}
\end{table}

%% file: appendix/C_full_peft.tex

\section{Full Low-Rank (PEFT) Results}
\label{app:peft_full}

Table~\ref{tab:peft_full} provides the complete PEFT results for the COGS with translation preservation setting, including methods at different parameter scales.

\begin{table}[htbp]
\centering
\caption{Complete PEFT results on COGS with translation preservation. Methods are grouped by trainable parameter count. 5 seeds where available; $^\dagger$ denotes 3 seeds. Translation baseline: chrF = $0.413 \pm 0.010$, PPL = $4.35 \pm 0.19$.}
\label{tab:peft_full}
\small
\begin{tabular}{@{}lrrrrr@{}}
\toprule
\textbf{Method} & \textbf{Params} & \textbf{COGS EM $\uparrow$} & \textbf{chrF $\uparrow$} & \textbf{PPL $\downarrow$} & \textbf{Best Ep} \\
\midrule
\multicolumn{6}{c}{\textit{$\sim$3M parameters}} \\
\midrule
PiSSA-r3$^\dagger$      & 3.27M & $98.60_{\pm 0.80}$ & $0.390_{\pm 0.009}$ & $6.31_{\pm 0.40}$ & 5.7 \\
DoRA-r3$^\dagger$       & 3.84M & $98.00_{\pm 1.25}$ & $0.408_{\pm 0.016}$ & $4.93_{\pm 0.20}$ & 5.7 \\
LoRA-r3$^\dagger$       & 3.27M & $97.53_{\pm 1.21}$ & $0.406_{\pm 0.015}$ & $4.95_{\pm 0.20}$ & 6.3 \\
\midrule
\multicolumn{6}{c}{\textit{$\sim$17M parameters (main comparison)}} \\
\midrule
LoRA-r16 replay10  & 17.43M & $\mathbf{95.84_{\pm 1.77}}$ & $0.405_{\pm 0.010}$ & $4.55_{\pm 0.17}$ & 10.0 \\
LoRA-r16 replay1   & 17.43M & $95.52_{\pm 1.78}$ & $0.414_{\pm 0.009}$ & $4.50_{\pm 0.19}$ & 9.6 \\
LoRA-r16           & 17.43M & $95.44_{\pm 1.75}$ & $0.414_{\pm 0.010}$ & $4.54_{\pm 0.20}$ & 9.6 \\
LoRA-r16 replay5   & 17.43M & $95.40_{\pm 1.65}$ & $0.410_{\pm 0.008}$ & $4.45_{\pm 0.20}$ & 9.0 \\
\FProjlr$^\dagger$ & 17.43M & $94.60_{\pm 2.29}$ & $0.413_{\pm 0.010}$ & $\mathbf{4.38_{\pm 0.18}}$ & 10.8 \\
\Foralr$^\dagger$ & 17.76M & $94.48_{\pm 2.21}$ & $0.413_{\pm 0.009}$ & $\mathbf{4.38_{\pm 0.19}}$ & 11.0 \\
OPLoRA-r16         & 17.43M & $94.44_{\pm 2.55}$ & $0.414_{\pm 0.010}$ & $4.54_{\pm 0.18}$ & 10.8 \\
CorDA-k16-r16      & 17.43M & $94.16_{\pm 1.40}$ & $0.414_{\pm 0.009}$ & $\mathbf{4.38_{\pm 0.19}}$ & 8.8 \\
\midrule
\multicolumn{6}{c}{\textit{Larger parameter scales}} \\
\midrule
LoRA-r64            & 69.73M & $97.64_{\pm 0.95}$ & $0.410_{\pm 0.011}$ & $4.82_{\pm 0.21}$ & 5.6 \\
DoRA-r16$^\dagger$  & 34.09M & $96.90_{\pm 0.42}$ & $0.418_{\pm 0.001}$ & $4.38_{\pm 0.06}$ & 10.0 \\
\bottomrule
\end{tabular}
\end{table}

Key observations from the full PEFT results:
\begin{enumerate}[leftmargin=*,nosep]
\item \textbf{Replay dominates COGS EM} at matched parameters (top-3 methods are replay-based), but at the cost of weaker PPL ($4.45$--$4.55$).
\item \textbf{Function-space methods achieve best PPL} ($4.38$, matching the translation baseline of $4.35$), indicating superior capability preservation.
\item \textbf{CorDA and our method converge on similar PPL} ($4.38$), supporting the shared insight that activation-derived subspaces protect capabilities.
\item \textbf{Increasing rank does not monotonically improve performance}: LoRA-r64 ($69.73$M) achieves $97.64\%$ COGS but with much worse PPL ($4.82$) than 17M-parameter methods.
\end{enumerate}

\begin{figure}[htbp]
    \centering
    \includegraphics[width=0.85\textwidth]{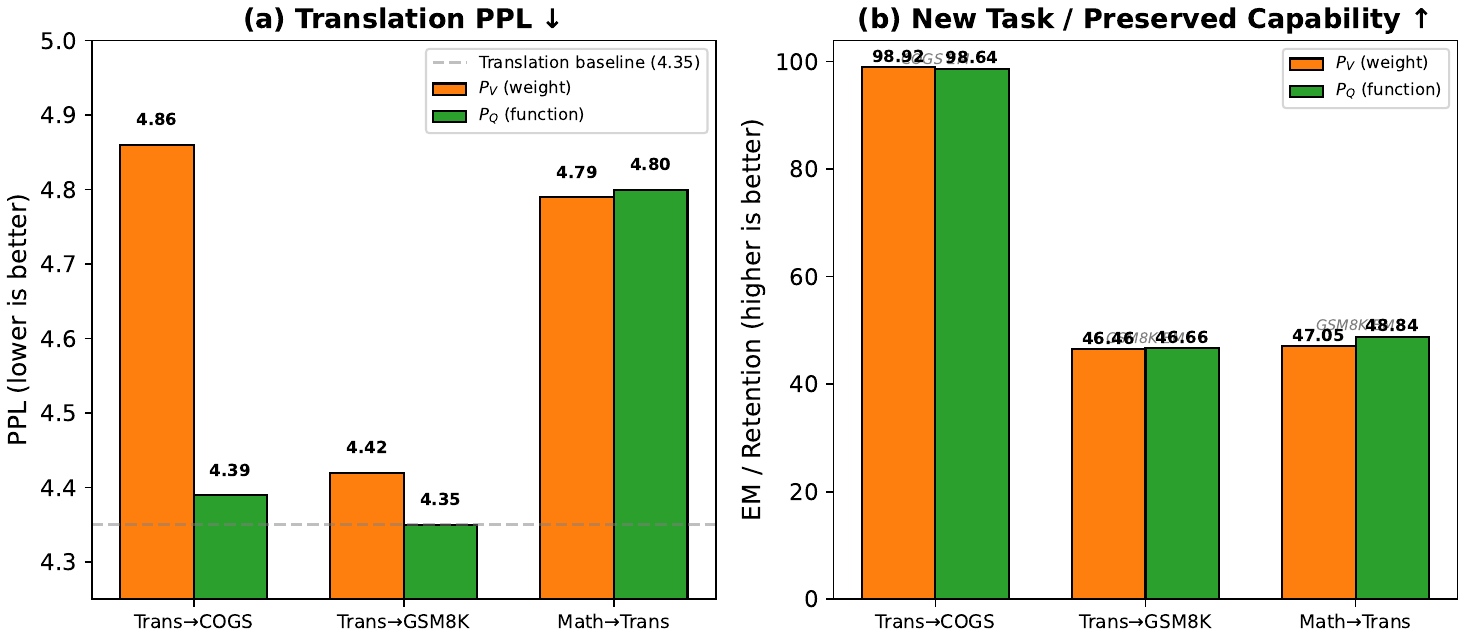}
    \caption{\textbf{Projection source across three settings} (appendix visualization of Table~\ref{tab:proj_source}). (a) Translation PPL: $\PQ$ (function-space) is consistently closer to baseline than $\PV$ (weight-space). (b) New-task or preserved metric: $\PQ$ is comparable or better. The systematic advantage of $\PQ$ confirms that the protected subspace's \emph{source} is what matters.}
    \label{fig:proj_source}
\end{figure}

%% file: appendix/D_spectral_only.tex

\section{Spectral-Only Baseline}
\label{app:spectral_only}

To clarify the role of the spectral calibration channel $\bm{\delta}$, we report an experiment where \emph{only} the diagonal calibration branch is trained: $\dW = \U_2\D_\delta\V_2^\top$ (no $\M$ or $\B\A$ branch).
This isolates what the channel can do on its own---establishing it as a genuine, low-degree-of-freedom plasticity channel rather than a no-op---and shows why it is \emph{paired with}, not substituted for, the high-capacity protected branch.
It is compared against no fine-tuning and standard LoRA-r16 on the SVAMP~\citep{patel2021svamp} mathematical reasoning task, with GSM8K retention as the preserved capability.

\begin{table}[htbp]
\centering
\caption{Spectral-only baseline on SVAMP fine-tuning with GSM8K retention. Model: Qwen3-1.7B (GSM8K LoRA merged).}
\label{tab:spectral_only}
\begin{tabular}{@{}lrrrr@{}}
\toprule
\textbf{Method} & \textbf{SVAMP $\uparrow$} & \textbf{GSM8K Retention $\uparrow$} & \textbf{Epoch} & \textbf{Val Loss} \\
\midrule
No fine-tuning  & 70.7 & 49.7 & --- & 2.87 \\
Diagonal-only   & $72.7_{\pm 4.7}$ & $23.7_{\pm 0.6}$ & 25 & 0.057 \\
LoRA-r16        & $70.7_{\pm 2.4}$ & $20.4_{\pm 2.4}$ & 8  & 0.044 \\
\bottomrule
\end{tabular}
\end{table}

The diagonal-only configuration (Table~\ref{tab:spectral_only}) achieves modest new-task learning (SVAMP $72.7\%$ vs.\ baseline $70.7\%$) and some GSM8K retention ($23.7\%$)---better than LoRA-r16 ($20.4\%$), but far below the baseline ($49.7\%$).
This shows that:
\begin{enumerate}[leftmargin=*,nosep]
\item The spectral calibration channel alone carries \emph{non-zero}, real plasticity---enough for minor adaptation, but not enough to learn a full task by itself.
\item It is not designed to preserve a capability on its own: without the projectors, GSM8K retention still drops from $49.7\%$ to $23.7\%$.
\item In the full method the projectors ($\PU$, $\PQ$) supply the structural protection while $\bm{\delta}$ contributes the bounded, controlled plasticity that Proposition~\ref{prop:controlled} reserves for the protected directions---two roles that compose rather than compete.
\end{enumerate}

%% file: appendix/E_extensions.tex

\section{Sensitivity to Subspace Dimensions, and Planned Extensions}
\label{app:extensions}

\subsection{\texorpdfstring{Sensitivity to the subspace dimensions $r$ and $k_f$}{Sensitivity to the subspace dimensions r and k-f}}
\label{app:dim_sweep}

\Fora{} has two structural hyperparameters: the SVD frozen rank $r$ in $\PU$ (default $100$) and the function-subspace size $k_f$ in $\PQ$ (default $16$). We sweep each independently on the Math$\to$Trans setting (preserving GSM8K reasoning while learning translation), holding the other at its default and running $5$ seeds per configuration. The default $(r,k_f)=(100,16)$ row is the \Fora{} result already reported in Table~\ref{tab:math_pres}. Table~\ref{tab:dim_sweep} collects both sweeps.

\begin{table}[htbp]
\centering\small
\caption{\textbf{Sensitivity to the subspace dimensions} on Math$\to$Trans (preserve GSM8K, learn translation); 5 seeds per row. Left: vary the SVD frozen rank $r$ at $k_f=16$. Right: vary the function-subspace size $k_f$ at $r=100$. The $(r,k_f)=(100,16)$ row is the default \Fora{} configuration (Table~\ref{tab:math_pres}). Across both sweeps GSM8K retention stays within a $0.4$-point band and translation chrF/PPL are essentially unchanged.}
\label{tab:dim_sweep}
\begin{tabular}{@{}lrrr@{\hskip 2.2em}lrrr@{}}
\toprule
\multicolumn{4}{c}{\textbf{Frozen rank $r$ ($k_f=16$)}} & \multicolumn{4}{c}{\textbf{Function size $k_f$ ($r=100$)}} \\
\cmidrule(r){1-4}\cmidrule(l){5-8}
$r$ & \textbf{GSM8K} & \textbf{chrF} & \textbf{PPL} & $k_f$ & \textbf{GSM8K} & \textbf{chrF} & \textbf{PPL} \\
\midrule
$20$        & $48.61_{\pm 0.64}$ & $0.390$ & $4.80$ & $4$         & $48.63_{\pm 0.25}$ & $0.389$ & $4.80$ \\
$50$        & $48.82_{\pm 0.33}$ & $0.390$ & $4.80$ & $8$         & $49.01_{\pm 0.35}$ & $0.389$ & $4.80$ \\
$100$ (def.) & $48.84_{\pm 0.42}$ & $0.390$ & $4.80$ & $16$ (def.) & $48.84_{\pm 0.42}$ & $0.390$ & $4.80$ \\
$200$       & $48.79_{\pm 0.34}$ & $0.390$ & $4.80$ & $32$        & $48.66_{\pm 0.37}$ & $0.390$ & $4.80$ \\
$400$       & $48.95_{\pm 0.39}$ & $0.390$ & $4.81$ & $64$        & $48.95_{\pm 0.25}$ & $0.390$ & $4.80$ \\
\bottomrule
\end{tabular}
\end{table}

Both hyperparameters are inconsequential within the ranges tested. Varying $r$ over a twentyfold range ($20\!\to\!400$) moves GSM8K retention by $0.34$ points ($48.61$--$48.95$) and leaves translation chrF and PPL unchanged; varying $k_f$ over a sixteenfold range ($4\!\to\!64$) moves it by $0.38$ points ($48.63$--$49.01$). Every configuration lies within one standard deviation of the default, so the differences are not statistically meaningful and the default $(100,16)$---fixed a priori---is not a tuned operating point. The behavior is consistent with the structural account of the method: once $\Q$ spans the dominant activation directions of the preserved capability, enlarging it adds near-degenerate directions that the high-capacity branch was already largely avoiding, and enlarging $r$ similarly saturates the output-side prior.

We emphasize that this robustness concerns \emph{how many} directions are protected, not \emph{which subspace} they are drawn from. The latter is the load-bearing design choice, and the projection-source ablation of Table~\ref{tab:proj_source} shows it is decisive: replacing the capability-derived $\PQ$ by the weight-derived $\PV$ degrades the preserved metric in every setting, even though both project away a subspace of comparable rank. Insensitivity to $k_f$ therefore makes \Fora{} easy to deploy without per-task tuning; it does not weaken the central claim that the projection source is what matters.

\subsection{Planned Extensions for Future Versions}

We outline additional analyses planned for future versions of this manuscript.

\paragraph{\texorpdfstring{Random $Q$ baseline.}{Random Q baseline.}}
To further verify that the source of $\Q$ matters, we plan to compare the learned $\Q$ (from preserved-capability activations) against a random orthogonal projector of the same rank. If random $\Q$ fails to protect capabilities, it strengthens the claim that activation-derived subspaces carry capability-specific information.

\paragraph{Calibration sample size sensitivity.}
We plan to evaluate preservation quality as a function of calibration sample count $N \in \{50, 100, 250, 500, 1000\}$, measuring whether $\Q$ stabilizes with modest data and whether domain shift in calibration inputs affects protection quality.

\paragraph{Multi-capability union projection.}
For simultaneous protection of multiple capabilities (e.g., translation $+$ math $+$ code), we plan to construct $\Q_{\text{union}} = [\Q_1 \; \Q_2 \; \Q_3]$ and evaluate the resulting trade-off between protection breadth and adaptive capacity.

\paragraph{Full-rank replay comparison.}
A direct comparison of our full-rank method against full fine-tuning with data replay (mixing $1\%$, $5\%$, $10\%$ of preserved-capability data) would quantify the complementarity (or redundancy) of structural versus data-level protection.

\paragraph{Larger-model validation.}
Scaling experiments to Qwen3-4B and Qwen3-8B, and to models from other families (e.g., LLaMA), would test whether the relative advantage of function-space over weight-space protection grows, shrinks, or remains constant with model scale.